\documentclass[fleqn,10pt]{wlscirep}
\usepackage[utf8]{inputenc}
\usepackage[T1]{fontenc}
\usepackage{cite}
\usepackage{url}
\author[1,*]{Mona Minakshi}
\author[2]{Pratool Bharti}
\author[1]{Tanvir Bhuiyan}
\author[1]{Sherzod Kariev}
\author[1,+]{Sriram Chellappan}
\affil[1]{Dept. of Computer Science and Engineering, University of South Florida, Tampa, FL 33620, USA}
\affil[2]{Dept. of Computer Science, Northern Illinois University, Dekalb, IL 60115 USA}

\affil[*]{mona6@usf.edu}
\affil[+]{ sriramc@usf.edu}
\usepackage{booktabs} % For formal tables

\usepackage{setspace}
\usepackage{caption}
\usepackage{comment}
\usepackage{xcolor}
\usepackage{amsmath}
\usepackage{comment}

\usepackage{balance}
 \usepackage{graphicx} 

\usepackage{tabularx}

\usepackage[noend]{algorithmic}
\usepackage{algorithmic}
\usepackage{newtxmath}

\usepackage{textcomp}

\usepackage[ruled]{algorithm2e}

\usepackage{setspace}

\usepackage{array}
\usepackage{mdwmath}
\usepackage{mdwtab}
\usepackage{booktabs}
\usepackage{multicol}
\usepackage{multirow}
\usepackage{rotating}
\usepackage{tabularx}
\usepackage[stable]{footmisc}
\usepackage[english]{babel}

\newcolumntype{L}[1]{>{\raggedright\let\newline\\\arraybackslash\hspace{1pt}}m{#1}}
\newcolumntype{C}[1]{>{\centering\let\newline\\\arraybackslash\hspace{1pt}}m{#1}}
\newcolumntype{R}[1]{>{\raggedleft\let\newline\\\arraybackslash\hspace{1pt}}m{#1}}
\title{A Framework based on Deep Neural Networks to Extract Anatomy of Mosquitoes from Images}

\begin{abstract}
\vspace{-5mm}
We design a framework based on  Mask  Region-based Convolutional Neural Network (Mask R-CNN) to automatically detect and separately extract anatomical components of mosquitoes - thorax, wings, abdomen and legs from  images. Our training   dataset consisted of $1500$ smartphone images of nine mosquito  species trapped in  Florida. In the proposed technique, the first  step is to detect anatomical components within a mosquito image. Then, we localize and classify the extracted anatomical components, while simultaneously adding a branch in the neural network architecture  to segment pixels containing only the anatomical components. Evaluation results are  favorable. To evaluate generality, we test our architecture trained only with mosquito images on  bumblebee images. We again reveal favorable results, particularly in extracting wings. Our techniques in this paper have practical applications in public health, taxonomy and citizen-science efforts.

 \end{abstract}
\begin{document}

\flushbottom
\maketitle

\section*{Introduction}
Mosquito-borne diseases  are still major public health concerns.  Across the world today, surveillance of mosquito vectors is still a manual process. Steps include trap placement,  collection of specimens, and identifying each specimen one by one under a microscope to determine the genus and species. Unfortunately, this process  is cognitively demanding and takes hours to complete. This is because, mosquitoes that fall into traps include both vectors, and also many that are not vectors. Recently, AI approaches are being designed to automate  classification of mosquitoes. Works like \cite{minakshi2018leveraging,de2016detection,fuchida2017vision,favret2016machine}  design machine learning models based on hand-crafted features from image data that are generated from either smartphones or digital cameras. Two recent $2020$ papers design deep neural network techniques (that do not need hand-crafted features) to classify mosquitoes from image data   generated via smartphones \cite{minakshi2020acm,park2020classification}. Other works process sounds of mosquito flight for classification, based on the notion that wing-beat frequencies are unique across  mosquito species \cite{chen2014flying,mukundarajan2016using,vasconcelos2019locomobis,ravi2016preventive}. 

In this paper, we demonstrate novel applications for mosquito images when processed using AI techniques  Since, the most descriptive anatomical components of mosquitoes are the thorax, abdomen, wings and legs, we present a technique in this paper that  extracts just the pixels comprising of these specific anatomical components from any mosquito image. Our technique is based on Mask Region-based Convolutional Neural Network\cite{he2017mask}. Here, we first extract  feature maps from our training dataset of $1500$ smartphone images of $200$ mosquito specimens spread across nine species trapped in Florida. Our network to extract feature maps is ResNet-101 with a Feature Pyramid Network \cite{lin2017feature} (an architecture that can handle images at multiple scales, and one well suited for our problem). Subsequently, we  detect and localize anatomical components only (denoted as foreground) in the images in the form of rectangular anchors. Once the foreground is detected, the next step is to segment the foreground pixels by adding a branch to mask (i.e., extract pixels of) each component present in the foreground. This is done in parallel with two other  branches to classify the extracted rectangular anchors and to tighten them to improve accuracy. Evaluation of our technique reveals  favorable results. We see that the thorax, wings, abdomen and legs are extracted with high  Precision (i.e., very low False Positives). For  legs though, False Negatives are high, since the number of background pixels overwhelm the number of leg pixels in the image. Nevertheless, we see that enough descriptive features within the leg of a mosquito are indeed extracted out, since mosquito legs are  long, and the descriptive features do repeat across the leg. 

We believe that extracting images of mosquito anatomy has  impact towards a) faster classification of mosquitoes in the wild; b) new digital-based, larger-scale and low-cost training programs for taxonomists; c) new and engaging tools to stimulate broader participation in citizen-science efforts and more. Also, to evaluate generality,  we tested our architecture trained on mosquito images with images of bumblebees (which are important pollinators). We see excellent accuracy in extracting the wings, and to a certain extent, the thorax, hence demonstrating the generality of our technique for many classes of insects.

\section*{Results}

\begin{figure}[ht]
\centering
%\hspace*{.9em}
    	\begin{minipage}[b]{\textwidth}
    	\includegraphics[width=\textwidth,height=9cm]{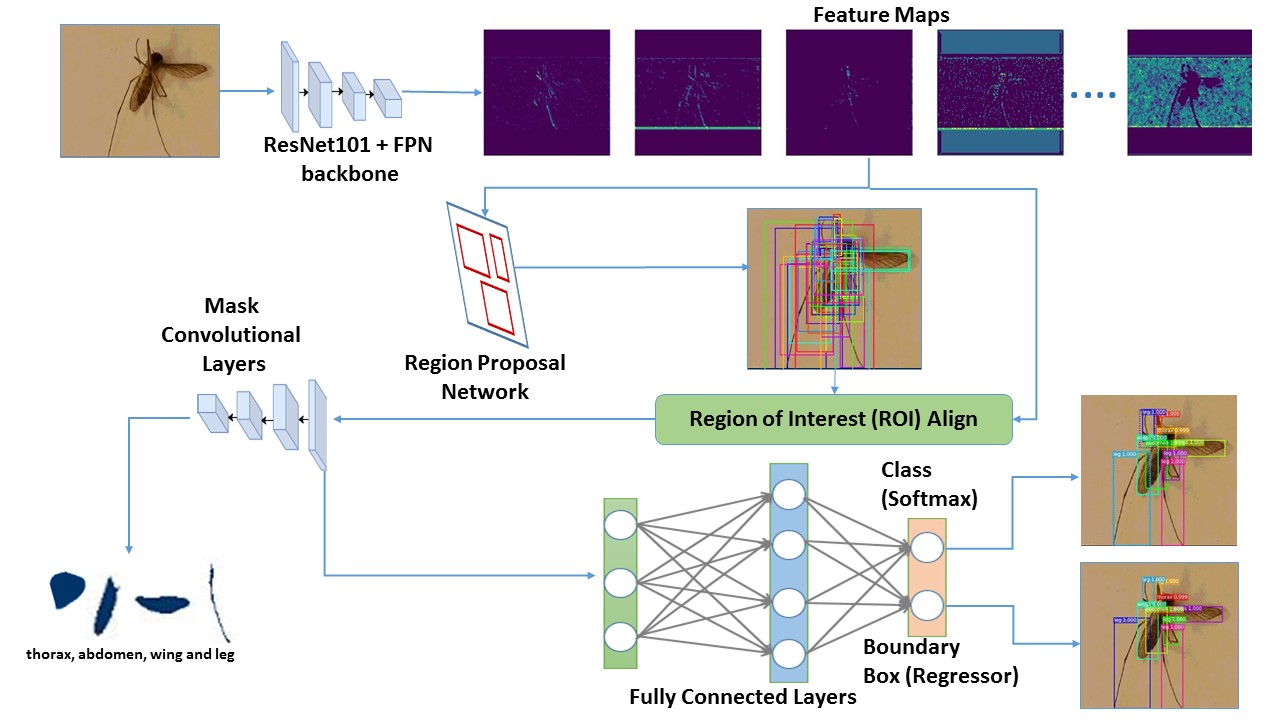}
	\end{minipage} 

\caption {The workflow of our architecture based on Mask R-CNN}
\label{fig:maskrcnnrachitecture}

\end{figure}

We trained a Mask Region-based Convolutional Neural Network (Mask R-CNN) to automatically detect and separately extract anatomical components of mosquitoes - thorax, wings, abdomen and legs from  images. For this study, we utilized  $23$ specimens of {\em Aedes aegypti} and {\em Aedes infirmatus}, and $22$ specimens of {\em Aedes taeniorhynchus}, {\em Anopheles crucians}, {\em Anopheles quadrimaculatus}, {\em Anopheles stephensi}, {\em Culex coronator}, {\em Culex nigripalpus} and {\em Culex salinarius}. After imaging the specimens via multiple smartphones, our dataset  was $1600$ mosquito images. These were split into $1500$ images for training the neural network, and $100$ images for validation. Together, this dataset yielded $1600$ images of thorax, $1600$ images of abdomen, $3109$ images of wings and $6223$ images of legs.  We trained our architecture illustrated in Figure \ref{fig:maskrcnnrachitecture} on an Nvidia graphic processing unit (GPU) cluster of four GeForce GTX TITAN X  cards having  $3,583$ cores and $12$ GB memory each. It took  $48$ hours to train and validate the architecture. For testing, we trapped and imaged (via smartphones) another set of $27$ mosquitoes, i.e., three per species. The testing data set consisted of $27$ images of thorax and abdomen, $48$ images of wings and $105$ images of legs.

%To train our mask-rcnn model, we used $1500$ mosquito images for training and $100$ for validation. We kept away $27$ mosquito images in test to further evaluate the model once training and fine-tuning is complete. Test dataset includes $27$ thorax, $27$ abdomen, $48$ wings and  $105$ legs. 

\begin{figure}[ht]
\centering
\hspace*{.9em}
    %\vspace{-2.0em}
    	\begin{minipage}[b]{.95\textwidth}
    	\includegraphics[width=\textwidth,height=7.5cm]{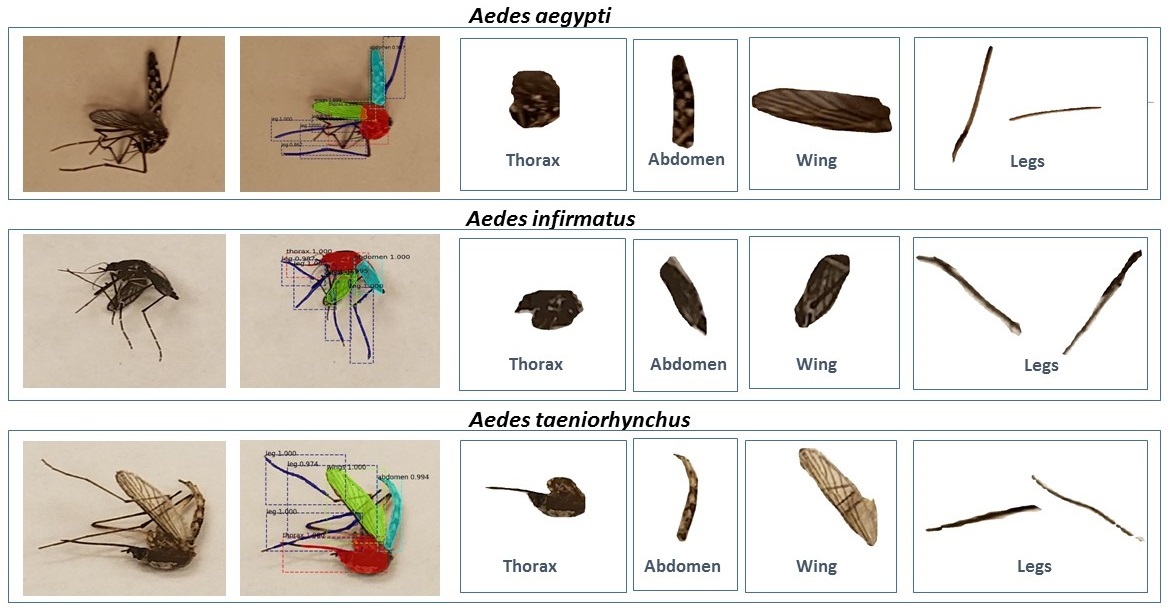}
	\end{minipage} \hspace*{0.9em}
	\begin{minipage}[b]{.95\textwidth}
    	\includegraphics[width=\textwidth,height=7.5cm]{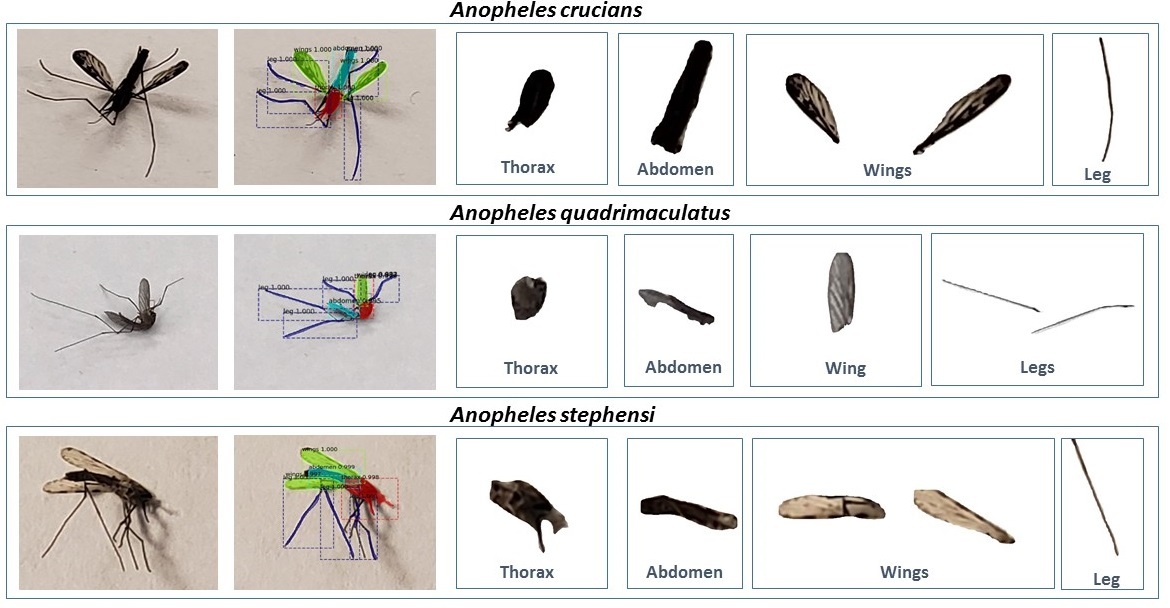}
	\end{minipage} \hspace*{0.9em}
	\begin{minipage}[b]{.95\textwidth}
    	\includegraphics[width=\textwidth,height=7.5cm]{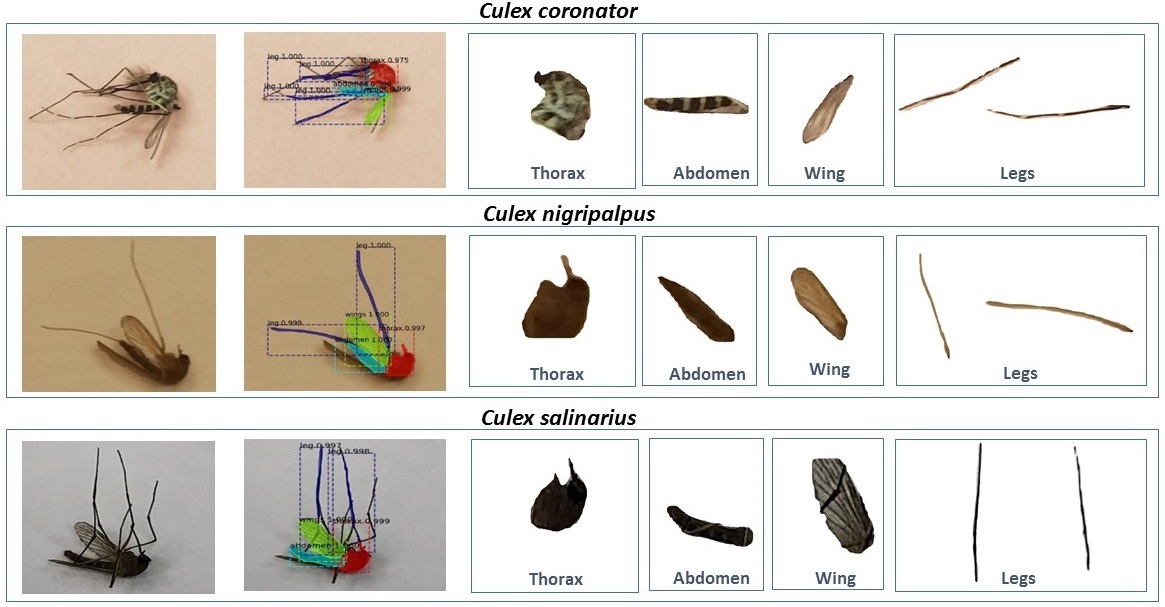}
	\end{minipage} \hspace*{.9em}

\caption {Results of extracting anatomical components for one sample image among the nine mosquito species in our dataset}
\label{fig:segmented}

\end{figure}

First, we visually present results of our technique to extract anatomical components of a mosquito in Figure \ref{fig:segmented} for one sample image among the nine species in our testing dataset. This figure is representative of all other images tested. We  see that the anatomical components are indeed coming out clearly from image data. Next, we quantify performance for our entire dataset using four standard metrics: Precision, Recall,  Intersection over Union (IoU) and Mean Average Precision (mAP). Precision is basically the fraction of relevant instances (here, pixels) among those instances (again, pixels) that are retrieved. Recall is the fraction of the relevant instances that were actually retrieved.   IoU is a  metric that assesses the ratio of areas of the intersection and the union among the predicted pixels and the ground truth.  A higher IoU means more overlap between predictions and the ground-truth, and so better classification. To define our final metric, the Mean Average Precision (mAP), we define another metric, Average precision (AP), which is the average of all the Precision values for a  range of Recall ($0$ to $100$ for our problem) at a certain preset IoU threshold for a particular class among the four for our problem (i.e., wings, thorax, legs and abdomen). This metric essentially balances both Precision and Recall for a particular value of IoU  for one class. Finally, the Mean Average Precision (mAP) is the average of  AP values among all our four classes.

The Precision and Recall values for the validation and testing datasets are presented in 
%Figures \ref{fig:presRecall} 
Tables \ref{table:presRecallValidation} and \ref{table:presRecallTesting} respectively for various values of IoU. We see that the performance metrics in the validation dataset during training  match the metrics during testing (i.e., unseen images) post training across all IoUs. This convinces us that our architecture is robust and not overfitted.  Precision for all classes is  high, which means that false positives are  low. Recall is  also high for the thorax, abdomen and wings, indicating low false negatives for these classes. However,  Recall for  legs class is relatively poor. It turns out that  a non-trivial portion of the leg pixels are  classified as the background in our architecture. While this may seem a bit discouraging, we once again direct readers to Figure \ref{fig:segmented}, wherein we can see that a very good portion of the legs are still identified and extracted correctly by our architecture (due to the high Precision). As such, the goal of gleaning the morphological markers from all anatomical components is still enabled. Finally, the mean Average Precision is presented in %Figure \ref{fig:apScores} 
Table \ref{table:apScoring} for all classes. The lower numbers in %Figure \ref{fig:apScores}
Table \ref{table:apScoring}, are due to poorer performance for classifying legs, as compared to thorax, abdomen and wings.

%Figure \ref{fig:apScores} shows the Average Precision score (as defined earlier) for all classes over all Recall values  for validation and testing dataset  at IoU thresholds of $0.30$, $0.50$ and $0.70$. We can see the Average Precision is indeed high for Thorax, Abdomen and Wings, but poor for legs. In Figure PP, we show the Mean Average Precision (mAP) of all four classes for validation and testing datsets, and naturally, the relatively low Precision values for the legs, results in a slighly lower mAP value

\begin{figure}[ht]
\centering
%\hspace*{.9em}
    %\vspace{-2.0em}
    	\begin{minipage}[b]{.95\textwidth}
    	\includegraphics[width=\textwidth,height=3.5cm]{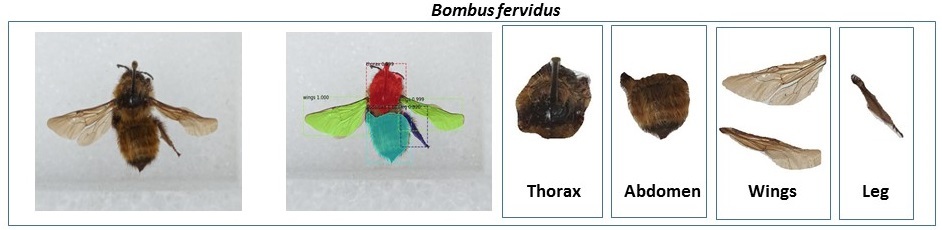}
	\end{minipage}
	\begin{minipage}[b]{.95\textwidth}
    	\includegraphics[width=\textwidth,height=3.5cm]{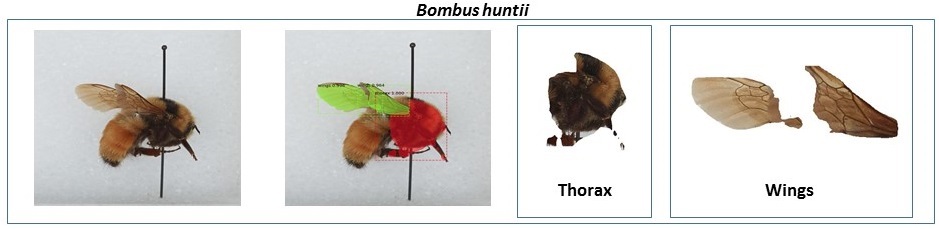}
	\end{minipage} 
	\begin{minipage}[b]{.95\textwidth}
    	\includegraphics[width=\textwidth,height=3.5cm]{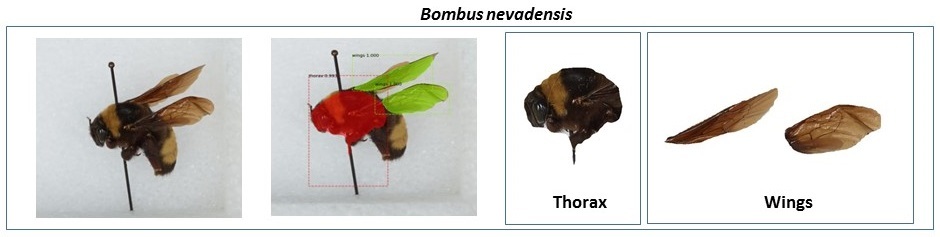}
	\end{minipage} 

\caption {Results of extracting anatomical components for bumblebees\cite{nmnh,nmnh1,nmnh2}}
\label{fig:segmented-bees}

\end{figure}

{\bf Results from a small experiment with bumblebee images:} We subsequently verified how our AI architecture that was trained only with mosquito images, performs, when tested with images of bumblebees. Bumblebees (Genus: {\em Bombus}) are  important pollinators, and detecting them in nature is vital. Figure \ref{fig:segmented-bees} presents our results for one representative image among three species of bumblebees, although our results are representative of more than $100$ bumblebee images we tested. Our image source for bumblebees was Smithsonian National Museum of Natural History in Washington, D.C. Images can be found here\cite{smithsonianinstitution}. As we  see in Figure \ref{fig:segmented-bees}, our technique is robust in detecting and extracting wings. While the thorax is mostly extracted correctly, the ability to extract out the abdomen and legs is relatively poor. With these results, we are  confident that our architecture in its present form, could be used to extract wings of many insects. With appropriate ground-truth data, only minimal tweaks to our architecture will be needed to ensure robust extraction of all anatomical components for a wide range of insects.

%There is certainly a high degree of confusion between the thorax and abdomen, but one we believe can be overcome with sufficient ground truth data and minimal tweaks to our architecture. As such, we believe that our technique is indeed generalizable for arthropods across the board, and is part of our future work

%In order to evaluate our technique with another arthropod of interest (in the much happier domain of pollination), we looked at how our architecture designed specifically for mosquitoes, works with images of bumblebees. Briefly, bumblebees (Genus: {\em Bombus}) are important pollinators in many parts of the world, with significant efforts being expended for their conservation, and their morphologies are also of interest to conservationists for identification (especially to contrast them with morphologies of more invasive species like honeybees). Figure \ref{fig:segmented-bees} presents our results for one representative image among four species of bumblebees \footnote{Image source was }. As we can see, our technique is very robust in detecting the wings. There is certainly a high degree of confusion between the thorax and abdomen, but one we believe can be overcome with sufficient ground truth data and minimal tweaks to our architecture. As such, we believe that our technique is indeed generalizable for arthropods across the board, and is part of our future work

\section*{Discussion}

We now present  discussions on the significance of contributions in this paper. 

{\bf a). Faster classification of trapped mosquitoes:} Across the world, where mosquito-borne diseases are problematic, it is standard practice to lay traps, and then come next day to pick up specimens, freeze them and bring them to a facility, where expert taxonomists identify each specimen one-by-one under a microscope to classify the genus and species. This process takes hours each day, and is cognitively demanding.  During rainy seasons and outbreaks, hundreds of mosquitoes get trapped, and it may take an entire day to process a batch from one trap alone. Based on technologies we design in this paper, we expect mobile cameras can assist in taking high quality pictures of trapped mosquito specimens, and the extracted anatomies can be used for classification by experts by looking at a digital monitor rather than peer through a microscope. This will result in lower cognitive stress for taxonomists and also speed up surveillance efforts. For interested readers, Table \ref{table:species_detail} presents details on morphological markers that taxonomists look for to identify mosquitoes used in our study \cite{identifyingmosquito}.

%Also, accurate visual classification by humans requires complex expertise that takes many years to develop, and as such the average age of taxonomists that do this job is about $65$ years in Florida and Southern India, where we have partners. With a system like 

%Naturally, with these advance ages, the stress levels only increase. With a system like ours in place, it is much easier now for a mobile camera to hover over trapped mosquitoes, auto-focus and click a picture or two, upload images to a cloud where our AI techniques in this paper will extract anatomies out in real-time, and feed this information to a dashboard. The taxonomist can now simply look at digital images of anatomies, classify the specimen, and push a button to let the  camera move and take pictures of the next mosquito and the process repeats. Ten taxonomists that saw our results were convinced that the images we extracted possess rich morphological markers enough for classification. With our techniques in this paper, the need to peer through a microscope, look at each anatomical component of every mosquito visually, and manually log vector densities  will be avoided, as will the cognitive burden on senior citizens. These benefits also directly apply towards classification of other tiny arthropods.

{\bf b). AI and Cloud Support Education for Training Next-generation Taxonomists:} The process of training taxonomists today across the world consists of very few training institutes, which store a few frozen samples of local and non-local mosquitoes. Trainees interested in these programs are not only professional taxonomists, but also hobbyists. The associated costs to store frozen mosquitoes are not trivial (especially in low economy countries), which severely limit entry into these programs, and also make these programs expensive to enroll. With technologies like the ones we propose in this paper, digital support for trainees is enabled. Benefits include, potential for remote education, reduced operational costs of institutes, reduced costs of enrollment, and opportunities to enroll more trainees. These benefits when enabled in practice will have positive impact to taxonomy, entomology, public health and more.

{\bf c). Digital Preservation of Insect Anatomies under Extinction Threats:} Recently, there are concerning reports  that insects are disappearing at rapid rates. We believe that digital preservation of their morphologies could itself aid preservation, as more and more citizen-scientists explore nature and share data to identify species under immediate threat. Preservation of insect images many also help educate future scientists across a diverse spectrum.

%to see In the case of mosquitoes, the fossil record is so poor that it is not possible to establish the actual ages of the family and extant taxa. Edwards (1923) surmised that ‘The origin and phylogenetic history of the Culicidae must go back to well into the Mesozoic Era; and, from the small size and fragile nature of the insects, it is probably too much to hope that we can ever obtain much direct palaeontological evidence on these matters'

\section*{Conclusions} In this paper, we design a Deep Neural Network Framework to extract anatomical components - thorax, wings, abdomen and legs from mosquito images. Our technique is based on the notion of Mask R-CNN, wherein we learn feature maps from images, emplace anchors around foreground components, followed by classification and segmenting pixels corresponding to the anatomical components within anchors. Our results are  favorable, when interpreted in the context of being able to glean descriptive morphological markers for classifying mosquitoes.  We believe that our work in this paper has broader impact in public health, entomology, taxonomy education, and newer incentives to engage citizens in  participatory sensing.

\section*{Methods}
\subsection*{Generation of Image Dataset and Preprocessing}

In Summer $2019$, we partnered with Hillsborough county mosquito control district in Florida to lay  outdoor mosquito  traps over multiple days. Each  morning after laying traps, we  collected all captured mosquitoes, froze them in a portable container and took them to the  county lab, where taxonomists identified them for us. For this study, we utilized $23$ specimens of {\em Aedes aegypti} and {\em Aedes infirmatus}, and $22$ specimens of {\em Aedes taeniorhynchus}, {\em Anopheles crucians}, {\em Anopheles quadrimaculatus}, {\em Anopheles stephensi}, {\em Culex coronator}, {\em Culex nigripalpus} and {\em Culex salinarius}. We point out that specimens of eight species were trapped in the wild. The {\em An. stephensi} specimens alone were lab-raised  whose  ancestors were originally trapped in India. 

Each specimen was then emplaced on a plain flat surface, and then imaged using one smartphone (among iPhone $8$, $8$ Plus, and Samsung Galaxy $S8$, $S10$) in normal indoor light conditions. To take images, the smartphone was attached to a movable platform $4$ to $5$ inches  above the mosquito specimen, and three photos at different angles were taken. One directly above, and two at $45^{\circ}$ angles to the specimen opposite from each other. As a result of these procedures, we generated a total of $600$ images. Then,   $500$ of these images were preprocessed to generate the training dataset, and the remaining $100$ images were separated out for validation. For preprocessing, the images were scaled down to $1024 \times 1024$ pixels for faster training (which did not lower accuracy). The images were  augmented by adding Gaussian blur and randomly flipping them from left to right. These methods are standard in image processing, which better account for variances during run-time execution. After this procedure, our training  dataset increased to $1500$ images. 

Note here that all mosquitoes used in this study are vectors. Among these, {\em Aedes aegypti} is particularly dangerous, since it spreads Zika fever, dengue, chikungunya and yellow fever. This mosquito is also globally distributed now.

\subsection*{Our Deep Neural Network Framework based on Mask R-CNN} 

To address our goal of extracting anatomical components from a mosquito image, a straightforward approach is to try a mixture of Gaussian models to remove background from the image \cite{minakshi2018leveraging,stauffer1999adaptive}. But this will only remove the background, without being able to extract anatomical components in the foreground separately. There are other recent approaches in the realm also. One technique is U-Net\cite{ronneberger2015u}, wherein semantic segmentation based on  deep neural networks is proposed. However, this technique does not lend itself to instance segmentation (i.e., segmenting and labeling of pixels across multiple classes).  Multi-task Network Cascade\cite{dai2016instance} (MNC)  is an instance segmentation technique, but it is prone to information loss, and is not suitable for images as complex as mosquitoes with multiple anatomical components. Fully Convolutional Instance-aware Semantic Segmentation\cite{li2017fully} (FCIS) is another instance segmentation technique, but it is prone to systematic errors on overlapping instances and creates spurious edges, which are not desirable. DeepMask\cite{pinheiro2015learning} developed by Facebook, extracts masks (i.e., pixels) and then uses Fast R-CNN\cite{girshick2015fast} technique to classify the pixels within the mask. This technique though is slow as it does not enable segmentation and classification in parallel. Furthermore, it uses selective search to find out  regions of interest, which further adds to delays in training and inference.

%in compare to RoI Align used in our approach. RoIAlign has a large impact in improving mask accuracy by relative $10\%$ to $50\%$\cite{he2017mask}.
In our problem, we have leveraged Mask R-CNN\cite{he2017mask} neural network architecture for  extracting masks (i.e.,  pixels) comprising of objects of interest  within an image which eliminates  selective search, and also uses Regional Proposal Network (RPN)\cite{ren2015faster} to learn correct regions of interest. This approach best suited for quicker training and inference. Apart from that, it uses superior alignment techniques for  feature maps, which helps prevent information loss. The basic architecture is shown in Figure \ref{fig:maskrcnnrachitecture}. Adapting it for our problem requires a series of steps presented below.

%To do so, our proposed technique leverages the  Mask R-CNN neural network architecture as shown in Figure \ref{fig:maskrcnnrachitecture}, which is state-of-the-art in terms of extracting masks (or, pixels) comprising of objects of interest  within an image \cite{he2017mask}. Adapting this architecture for our problem requires a series of steps presented below.
\begin{itemize}

    \item {\bf Annotation for Ground-truth:} First, we manually annotate our training and validation images using VGG Image Annotator (VIA) tool \cite{dutta2019vgg}. To do so, we manually (and carefully) emplace bounding polygons around  each anatomical component in our training and validation images. The pixels within the polygons and associated labels (i.e., thorax, abdomen, wing or leg) serve as ground truth.  One sample annotated image is shown in Figure \ref{fig:annotated}.
    
    \item {\bf Generate Feature Maps using CNN:}  Then, we  learn semantically rich features in the training image dataset to recognize the complex anatomical components of the mosquito. To do so, our  neural network architecture is a combination of the popular Res-Net101 architecture with Feature Pyramid Networks (FPN)\cite{lin2017feature}. Very briefly, ResNet-101\cite{he2016deep} is a CNN with residual connections, and was specifically designed to remove vanishing gradients at later layers during training. It is relatively simple with $345$ layers. Addition of a feature pyramid network to ResNet was attempted in  another study, where the motivation was  to leverage the naturally pyramidal shape of CNNs, and to also create a subsequent feature pyramid network that combines low resolution semantically strong features with high resolution semantically weak features using a top-down pathway and lateral connections \cite{lin2017feature}. This resulting architecture is well suited to learn from images at different scales from only minimal input image scales. Ensuring scale-invariant learning is specifically important for our problem, since mosquito images can be generated at different scales during run-time, due to diversity in camera hardware and human induced variations. 
    
    %The total number of layers in this combined architecture is $359$, with $345$ convolutional layers and $14$ FPN layers.

    %The files are saved annotation details in a JSON file  as a set of $(x,y)$ coordinates. For faster training and overhead reduction,  we scaled down each image by $50\%$.
    \item{\bf Emplacing Anchors on Anatomical Components in the Image:} In this step, we leverage the notion of Regional Proposal Network (RPN)\cite{ren2015faster} and results from the previous two steps, to design a simpler CNN that will learn feature maps  corresponding to ground-truthed anatomical components in the training images. The end goal is to emplace anchors (rectangular boxes) that enclose the detected anatomical components of interest in the image.
    
    \item{\bf  Classification and Pixel-level Extraction:} Finally, we align the feature maps of the anchors (i.e., region of interest) learned from the above step into fixed sized feature maps which serve as input to three branches to: a) label the anchors with the anatomical component; b) extract only the pixels within the anchors that represents an anatomical component; and c) tighten the anchors for improved accuracy. All  three steps are done in parallel.
\end{itemize}

\subsection*{Loss Functions}

For our problem, recall that there are three specific sub-problems: labeling the anchors as thorax, abdomen, wings or leg;  masking the corresponding pixels within each anchor; and a regressor to tighten anchors. We elaborate now on the loss functions used for these three sub-problems. We do so because, loss functions are a critical component during training and validation of deep neural networks to improve learning accuracy and avoid overfitting.

%The first one is  correct classification of pixels in an image as belonging to an anatomical component of a mosquito (lets call it {\em pixel classification loss}), and the second is to label the classified pixels with the correct anatomical component (lets call it {\em label classification loss}). 

{\bf Labeling (or Classification) Loss:} For classifying the anchors, we utilized the  Categorical Cross Entropy loss function, and it worked well. For a single anchor $j$, the loss is given by,
\vspace{-1mm}
\begin{equation}
\label{eq:CCE}
CCE_j=-log(p),
%-\frac{1}{K}\displaystyle\sum_{i = 1}^{K}(\log(p)), 
\vspace{-0pt}
\end{equation}
\vspace{-4mm}

where   $p$ is the model estimated probability for the ground truth class of the anchor. 

%Naturally, for all  anchors $N$ in our training image dataset, the Categorical Cross Entropy Loss is given by

\begin{figure}[ht]
\centering
\hspace*{.9em}
        \begin{minipage}[b]{0.45\textwidth}
    	\includegraphics[width=\textwidth,height=5cm]{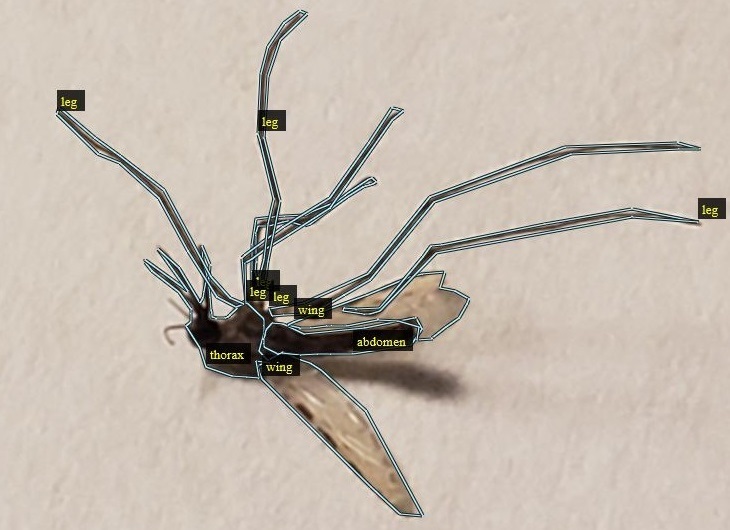} %with_focal_loss-2.JPG
    	\caption{Manual annotation of each anatomy (thorax, abdomen, wings, and legs) using VGG Image Annotator (VIA) tool}
    	\label{fig:annotated}
	\end{minipage}
    	\begin{minipage}[b]{0.45\textwidth}
    	\includegraphics[width=\textwidth,height=5cm]{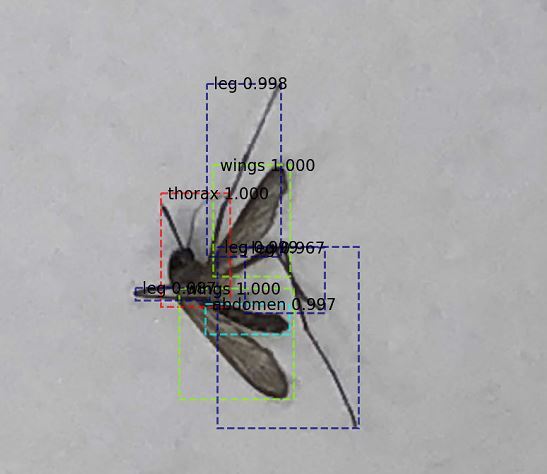}
    	\caption{After emplacement of anchors, we see significantly more background pixels than foreground pixels for anchors encompassing legs}
    	\label{fig:focalloss}
	\end{minipage}

%\caption {After emplacement of anchors, we see significantly more background pixels than foreground pixels for anchors encompassing legs}

\end{figure}

%where   $y_{i}$ is a $1$-d ground truth indicator if class $i$ (i.e., anatomical component) is true ($y_{i}=1$), or false ($y_{i}=0$); $K$ is the total number of  classes, which is four for our problem; and $p_{i}$ is the predicted probability of class $i$. Naturally, for all  anchors $N$ in our training image dataset, the Categorical Cross Entropy Loss is given by
\iffalse
\vspace{-4mm}
\begin{equation}
\label{eq:CCE_FULL}
CCE=-\frac{1}{N}\sum_{j = 1}^{N}CCE_j.
% \vspace{10pt}
\end{equation}
\vspace{-2mm}
\fi
{\bf Masking Loss:} Masking is most challenging, considering the  complexity in learning to detect only  pixels comprising of anatomical components in an anchor. Initially, we  experimented with the  simple  Binary Cross Entropy loss function. With this loss function, we noticed good accuracy for pixels corresponding to thorax, wings and abdomen. But, many pixels corresponding to  legs were mis-classified as background. This is because of class imbalance  highlighted in Figure \ref{fig:focalloss}, wherein we  see  significantly larger number of background pixels, compared to number of foreground pixels for anchors (colored blue) emplaced around legs. This  imbalance leads to poor learning for legs, because the binary class entropy loss function is biased towards the (much more, and easier to classify) background pixels. 

To fix this shortcoming, we investigated another more recent loss function called {\em  focal loss} \cite{lin2017focal} which lowers the effect of well classified samples on the loss, and rather places more emphasis on the  harder samples. This loss function hence prevents more commonly occurring background pixels from overwhelming the not so commonly occurring foreground pixels during learning, hence overcoming class imbalance problems. The focal loss for a pixel $i$ is represented as,
\vspace{-5mm}

\vspace{-5pt}
\begin{equation}
\label{eqn:focalloss}
FL(i)=-(1-p)^\gamma\log(p), 
\vspace{-0pt}
\end{equation}
%where recall that $p ~\epsilon ~ [0,1]$ is the estimated probability of our architecture for the class with denoting the ground-truth (or the anatomical component), and $\gamma \ge 0$ is a tunable parameter. The term $p_t$ is defined as, 

%{\bf write eq according to focal loss paper}
\vspace{-2mm}
%where $p_t$ is the larger value among the two probabilities outputted for a pixel as belonging to the foreground or background,
where $p$ is the model estimated probability for the ground truth class, and $\gamma$ is a tunable parameter, which was set as $2$ in our model. With these definitions, it is easy to see that when a pixel is mis-classified and $p \to 0$, then the modulating factor $(1-p)^\gamma$ tends to $1$ and the loss ($log(p)$) is not affected. However, when a pixel is classified correctly and when $p \to 1$, the loss is down-weighted. In this manner, priority during training is emphasized more on the hard negative classifications, hence yielding superior classification performance in the case of unbalanced datsets. Utilizing the focal loss gave us superior classification results for all anatomical components.

%The classification outcome with the focal loss for our architecture is shown in Figures [YY] (c) and also in Figure [yy] (d), where the leg portion is highlighted, and as we can see has much lower mis-classifications compared to mis-classifications in Figure [XX] (b).

%probability for the class label $y$=$1$, $p_t$ = $p$ or $1 - p$ depending on if the label is $1$ or $0$, respectively. In focal loss, we add a modulating factor $(1$-$ p_t)^\gamma$ to the above loss, having tunable focusing parameter    $\gamma$ >=0.

%In this case, when an anatomy is misclassified and $p_t$ is small, then the modulating factor is near to $1$ and then  there is no effect on loss. While $p_t$ tends, then modulating factor approaches $0$ and then loss of well classified cases gets down-weighted.  Effect on focal loss on final predicted images is shown in Figure \ref{fig:focalloss}(b)

{\bf Regressor Loss:} To tighten the anchors and hence improve masking accuracy, the loss function we utilized is based on the summation of  Smooth $L1$ functions computed across anchor, ground truth and predicted anchors. Let ($x,y$) denote the top-left coordinate of a predicted anchor. Let $x_a$ and $x^*$ denote the same for anchors generated by the RPN, and the manually generated ground-truth. The notations are the same for the $y$ coordinate, width $w$  and height $h$ of an anchor. We  define several terms first, following which the loss function $L_{reg}$ used in our architecture is presented.
\vspace{-2mm}
%top-left the box, and the logarithm of the heights and widths. The  smooth-L1 loss is represented as,

\iffalse
\begin{equation}
\begin{array}{l}
 L_{reg}(t_i,t_{i}^{*})=\sum_{i\epsilon{{x,y,w,h}}}smooth_{L_1}(t_{i}^{*}-t_i)\\ \\
 \textrm{in ~which,} \\

smooth_{L_1}= 
\begin{cases}
    0.5x^2 ,& \text{if } |x|< 1\\
    |x| -0.5,              & \text{otherwise}
\end{cases} \\ \\ 
\textrm{and,}\\ \\
t_x^*=\frac{(x^*-x_a)}{w_a},~~~~~~~~~~~~~~~~~~~~
t_y^*=\frac{(y^*-y_a)}{h_a},~~~~~~~~~~~~~~~~~~~~
t_w^*=\log(\frac{w^*}{w_a}),~~~~~~~~~~~~~~~~~~~~
t_h^*=\log(\frac{h^*}{h_a}),\\ \\
t_x=\frac{(x-x_a)}{w_a},~~~~~~~~~~~~~~~~~~~~~
t_y=\frac{(y-y_a)}{h_a},~~~~~~~~~~~~~~~~~~~~~
t_w=\log(\frac{w}{w_a}),~~~~~~~~~~~~~~~~~~~~
t_h=\log(\frac{h}{h_a})\\ \\
\end{array}
\vspace{-0pt}
\end{equation}
\fi
\begin{equation}
\begin{array}{l}

t_x^*=\frac{(x^*-x_a)}{w_a},~~~~~~~~~~~~~~~~~~~~
t_y^*=\frac{(y^*-y_a)}{h_a},~~~~~~~~~~~~~~~~~~~~
t_w^*=\log(\frac{w^*}{w_a}),~~~~~~~~~~~~~~~~~~~~
t_h^*=\log(\frac{h^*}{h_a}),\\ \\
t_x=\frac{(x-x_a)}{w_a},~~~~~~~~~~~~~~~~~~~~~
t_y=\frac{(y-y_a)}{h_a},~~~~~~~~~~~~~~~~~~~~~
t_w=\log(\frac{w}{w_a}),~~~~~~~~~~~~~~~~~~~~
t_h=\log(\frac{h}{h_a}),\\ \\
smooth_{L_1}= 
\begin{cases}
    0.5x^2 ,& \text{if } |x|< 1\\
    |x| -0.5,              & \text{otherwise}
\end{cases} ~~~\textrm{and} \\ \\ 
L_{reg}(t_i,t_{i}^{*})=\sum_{i\epsilon{{x,y,w,h}}}smooth_{L_1}(t_{i}^{*}-t_i).\\ \\
\end{array}
\vspace{-0pt}
\end{equation}
\vspace{-9mm}

\subsection*{Hyperparameters} 
For convenience, Table \ref{table:hyperparameter} lists  values of critical hyperparameters in our finalized architecture.

\subsection*{DATA AVAILABILITY}
The sample dataset is available at \href{https://github.com/mminakshi/Mosquito-Data/tree/master/Data}{https://github.com/mminakshi/Mosquito-Data/tree/master/Data}.
\subsection*{CODE AVAILABILITY}
We have leveraged \href{https://github.com/matterport/Mask\_RCNN}{Matterport Github repository} for Mask RCNN implementation. The code is open source and publicly available\cite{maskRCNN}. 

\bibliography{sample}

%% Here is the endmatter stuff: Supplementary Info, etc.
%% Use \item's to separate, default label is "Acknowledgements"
\section*{Acknowledgements}

We  acknowledge support from the Hillsborough County Mosquito Control District in helping us identify mosquitoes trapped in the wild.
%, and Dr. John Adams from the University of South Florida for providing samples of \textit{An. stephensi} mosquitoes. This work was supported in part by US National Science Foundation under Grant CBET \#1743985. Any opinions and findings are those of the authors alone and do not necessarily reflect the views of the funding agency.

\section*{Author contributions statement}
M. Minakshi led the design of Mask R-CNN algorithms and data collection for anatomy extraction. P. Bharti and M. Minakshi conducted experiments of various loss functions and fine tuned the neural network parameters. T. Bhuiyan conducted experiments on performance evaluations and collaborated with M. Minakshi in architecture design. S. Kariev assisted in data collection and annotation of ground-truth images. S. Chellappan conceptualized the project, and coordinated all technical discussions and choices of parameters for the neural network architectures.
%Must include all authors, identified by initials, for example:
%A.A. conceived the experiment(s),  A.A. and B.A. conducted the experiment(s), C.A. and D.A. analysed the results.  All authors reviewed the manuscript. 

\section*{Additional information}

We have no competing interests.

\begin{table}[ht]
\begin{center}
\centering \caption{Precision and Recall for different IoU thresholds on validation set}
\label{table:presRecallValidation}
\begin{tabular}{|l|l|l|l|l|l|l|}
\hline
\multirow{2}{*}{\textbf{Anatomy}} & \multicolumn{2}{l|}{\textbf{IoU Ratio=0.30}} & \multicolumn{2}{l|}{\textbf{IoU Ratio=0.50}} & \multicolumn{2}{l|}{\textbf{IoU Ratio=0.70}} \\ \cline{2-7} 
 & \textbf{Precision (\%)} & \textbf{Recall (\%)} & \textbf{Precision (\%)} & \textbf{Recall (\%)} & \textbf{Precision (\%)} & \textbf{Recall (\%)} \\ \hline
Thorax &    94.57       & 95.15       &  99.32         &   89.69     &  99.09         &  66.67      \\ \hline
Abdomen &     95.27      &    90.96    &   96.37        &  85.80      &  99.17         &  77.41      \\ \hline
Wing  &   98.17        & 91.49        &  98.53         &  85.50      &   97.82        &  76.59      \\ \hline
Leg &   99.35        &  37.85      &  100         &  25.60      &   100        &  21.50     \\ \hline
\end{tabular}
\end{center}
\end{table}
\begin{table}[ht]
\begin{center}

\centering \caption{Precision and Recall for different IoU thresholds on testing set}
\label{table:presRecallTesting}
\begin{tabular}{|l|l|l|l|l|l|l|}
\hline
\multirow{2}{*}{\textbf{Anatomy}} & \multicolumn{2}{l|}{\textbf{IoU Ratio=0.30}} & \multicolumn{2}{l|}{\textbf{IoU Ratio=0.50}} & \multicolumn{2}{l|}{\textbf{IoU Ratio=0.70}} \\ \cline{2-7} 
 & \textbf{Precision (\%)} & \textbf{Recall (\%)} & \textbf{Precision (\%)} & \textbf{Recall (\%)} & \textbf{Precision (\%)} & \textbf{Recall (\%)} \\ \hline
Thorax &     96      &    96    &   100        &  87.50      &   100        & 52       \\ \hline
Abdomen &  95.23         & 95.23       &   100        & 85.71       &  100         &  61.90      \\ \hline
Wing  &    100       &  88.36      & 100          &  81.81      &   100        & 61.36       \\ \hline
Leg &  95.46         &  35.76      & 100          &  21.40      &    100       &    19.25    \\ \hline
\end{tabular}
\end{center}
\end{table}
\begin{table}[ht]
\begin{center}
 \caption{mAP scores for masking}
	\label{table:apScoring}
\begin{tabular}{ |p{0.10\linewidth}| p{0.16\linewidth}| p{0.15\linewidth}| } 
 
 \hline
 \textbf{IoU Ratio} & \textbf{Validation set (\%)} & \textbf{Testing set (\%)} \\
 \hline
 
 $0.30$ & $62.50$ & $53.49$ \\ 
 \hline
$0.50$ & $60$ & $52.38$ \\ 
 \hline
$0.70$ & $51$ & $41.20$ \\ 
 \hline

\end{tabular}
\end{center}

\end{table}

\begin{table}[!ht]
\begin{center}
    
 \caption{Anatomical components and markers aiding mosquito classification\cite{ifas,glyshaw_wason,dharmasiri2017first,ifas1,floore1976anopheles}}
	\label{table:species_detail}
	%\cite{ifas}\cite{glyshaw_wason} \cite{dharmasiri2017first}\cite{ifas1}\cite{floore1976anopheles}
\begin{tabular}{ |p{4.2cm}| p{2.3cm}| p{2.3cm}| p{2.3cm}|p{2.3cm}| } 
 
 \hline
\textbf{Species} & \textbf{Thorax}  & \textbf{Abdomen}  & \textbf{Wing}  & \textbf{Leg} \\
 \hline
$Aedes~aegypti$  & dark with white lyre-shaped pattern and patches of white scales & dark with narrow white basal bands & dark & dark with white basal bands  \\
 \hline
 $Aedes~infirmatus$ & brown with patches of white scales &  dark with basal triangular patches of white scales & dark & dark \\
 \hline
  $Aedes~taeniorhynchus$ & dark with patches of white scales & dark with white basal bands & dark & dark with white basal bands \\
  \hline
   $ Anopheles ~crucians$ & gray-black  & dark & light and dark scales;  dark costa; white wing tip; 3 dark spots on sixth vein & dark with pale `knee' spots \\
  
    \hline
   $ Anopheles ~quadrimaculatus$ & gray-black & dark  & light and dark scales; 4 distinct darker spots & dark with pale `knee' spots \\
  \hline
  $ Anopheles ~stephensi$ & broad bands of white scales  &   & four dark spots on costa extending to first vein & speckling;  narrow white band on fifth tarsomere  \\
  \hline
  $ Culex ~coronator$ & dark with white
scales on the apical and  third
segments   & sterna without dark triangles; mostly pale scaled &  & distinct
basal and apical bands on hind tarsomeres  \\
  \hline
  $ Culex ~nigripalpus$ & brown copper color; white scales  & dark with lateral white patches & dark  & dark  \\
  \hline
$ Culex ~salinarius$ & copper; sometimes distinctly red; patches of white scales  & dark with golden basal bands; golden color on seventh segment & dark  & dark  \\
  \hline
\end{tabular}
\end{center}
\end{table}
%>{\centering}
\begin{table}[ht]
\begin{center}
 \caption{Values of critical hyperparameters in our architecture}
	\label{table:hyperparameter}
\begin{tabular}{ |p{0.20\linewidth}| p{0.20\linewidth}|  } 
 
 \hline
 \textbf{Hyperparameter} & \textbf{Value} \\
 \hline
 
 Number of Layers & $394$ \\ 
 \hline
 Learning Rate & 1e-3 for 1-100 epochs \newline 
                    5e-4 for 101-200 epochs \newline 1e-5 for 201-400 epochs \newline 1e-6 for 401-500 epochs \\ 
 \hline
Optimizer & SGD \\ 
 \hline
Momentum & $0.9$  \\ 
\hline
Weight Decay & $0.001$  \\ 
\hline
Number of Epochs & $500$  \\ 
\hline
\end{tabular}
\end{center}

\end{table}

\newpage

\iffalse
\begin{figure}[ht]
\centering
\hspace*{.9em}
    	\begin{minipage}[b]{0.45\textwidth}
    	\includegraphics[width=\textwidth,height=6cm]{images/Preciison_Recall_og Validation Set-4.JPG}
    	 \caption*{ a$)$. Validation Set }
	\end{minipage} \hspace*{.1em}
	\begin{minipage}[b]{0.45\textwidth}
    	\includegraphics[width=\textwidth,height=6cm]{images/Preciison_Recall_og Testing Set-4.JPG}
    	\caption*{ b$)$. Testing Set }
	\end{minipage}

\caption {Precision and Recall for different IoU thresholds on Validation and Testing Set}
\label{fig:presRecall}

\end{figure}
\fi
\iffalse
\begin{figure}[ht]
\centering
\hspace*{.9em}
    	\begin{minipage}[b]{0.65\textwidth}
    	\includegraphics[width=\textwidth,height=7cm]{images/AP_mask-6.JPG}
    	%\caption*{ b$)$.~Mask }
	\end{minipage} \hspace*{.1em}
\caption {mAP scores for Masking}
\label{fig:apScores}
\end{figure}
\fi

\iffalse
\begin{figure}[ht]
\centering
\hspace*{.9em}
    	\begin{minipage}[b]{0.70\textwidth}
    	\includegraphics[width=\textwidth,height=7.5cm]{images/image_annotation-1.jpg}
	\end{minipage} \hspace*{.1em}

\caption {Manual Annotation using VGG Image Annotator(VIA) tool}
\label{fig:annotated}

\end{figure}
\fi

%%
%% TABLES
%%
%% If there are any tables, put them here.
%%
\iffalse
\begin{table}[ht]
\centering
\caption{This is a table with scientific results.}
\medskip
\begin{tabular}{ccccc}
\hline
1 & 2 & 3 & 4 & 5\\
\hline
aaa & bbb & ccc & ddd & eee\\
aaaa & bbbb & cccc & dddd & eeee\\
aaaaa & bbbbb & ccccc & ddddd & eeeee\\
aaaaaa & bbbbbb & cccccc & dddddd & eeeeee\\
1.000 & 2.000 & 3.000 & 4.000 & 5.000\\
\hline
\end{tabular}
\end{table}
\fi
\end{document}